\documentclass[cameraready]{Interspeech}
\title{PyPhonPlan: Simulating phonetic planning with dynamic neural fields and task dynamics}

\author[orcid=0000-0003-4469-1512]{Sam}{Kirkham}

\address{
    Phonetics Laboratory, Lancaster University, United Kingdom
}

\email{s.kirkham@lancaster.ac.uk}

\keywords{computational modeling, phonetic planning, dynamic neural fields, task dynamics, speech production}

\begin{document}

\maketitle

\begin{abstract}
    We introduce PyPhonPlan, a Python toolkit for implementing dynamical models of phonetic planning using coupled dynamic neural fields and task dynamic simulations. The toolkit provides modular components for defining planning, perception and memory fields, as well as between-field coupling, gestural inputs, and using field activation profiles to solve tract variable trajectories. We illustrate the toolkit's capabilities through an example application:~simulating production/perception loops with a coupled memory field, which demonstrates the framework's ability to model interactive speech dynamics using representations that are temporally-principled, neurally-grounded, and phonetically-rich. PyPhonPlan is released as open-source software and contains executable examples to promote reproducibility, extensibility, and cumulative computational development for speech communication research.
\end{abstract}

\section{Introduction}

The production of speech appears at first to be unfathomably complex. Speakers must navigate the precise temporal coordination of a high-dimensional neuromuscular space with many degrees-of-freedom, with the ultimate goal of shaping a channel of moving air to achieve meaningful action in a rich social environment. Despite this apparent complexity, there is abundant evidence across biological motor control that organisms simplify the control of high-dimensional movement by constraining it to a smaller number of functionally relevant variables \cite{bernstein1967, haken-etal1985}. In the domain of speech production, one of the most popular models of these functionally relevant variables is task dynamics \cite{fowler1980, kelso-etal1986, saltzman-munhall1989, browman-goldstein1992, byrd-saltzman1998, iskarous2017}. Task dynamics hypothesizes that an abstract dynamical force called a \textit{gesture} drives low-dimensional \textit{tract variables} that control geometric properties of the vocal tract, such as lip aperture (LA) or tongue tip constriction degree (TTCD). Tract variables are a hypothesized coordinate system in which articulatory goals are defined, and are achieved by flexibly reconfiguring groups of articulators (tongue tip, lower lip, jaw) to meet the ongoing demands of speech.

A fundamental assumption in task dynamics is that tract variables are governed by point attractor dynamics, where a gesture drives a tract variable toward an equilibrium position. However, in conventional task dynamic models, these target values are stipulated as invariant parameters, with no intrinsic mechanism for learning, memory or sensory influences on speech production. A major development in this respect is the integration of dynamic neural fields for movement planning with task dynamic theory \cite{kirov-gafos2007, tilsen2007}. A dynamic neural field (DNF) is a low-dimensional representation of a functional neural population, which is defined over a metric variable for a sensory or movement parameter \cite{erlhagen-schoener2002, schoener-etal2016}. The field's dynamics evolve over time under the influence of sensory inputs, memory, and field-internal dynamics \cite{schneegans-etal2016, simmering-spencer2008}.

The last decade has produced a number of DNF models for speech planning \cite{kirov-gafos2007, tilsen2007, roon-gafos2016, iskarous2016, tilsen2019, harper2021, shaw-tang2023, stern-shaw2023, kirkham-strycharczuk2024, kirkham-etal2025, chaturvedi-shaw2025}. While much of this work has focused on relatively low-level phonetic variation, there is increasing evidence that DNFs can also model more traditionally `phonological' phenomena \cite{shaw2025}, as well as lexical meaning \cite{stern-pinango2026}. An excellent review of DNF models for speech communication can be found in \cite{stern2025}. In light of the current landscape, there is a need to develop modern computational frameworks that make this integration of dynamic neural fields and task dynamics accessible to a broader community. To address this gap, we introduce PyPhonPlan, a Python toolkit for simulating phonetic planning using coupled dynamic neural fields and task dynamic equations. The toolkit provides modular components for defining activation fields, coupling architectures, gestural inputs, and task dynamic simulations. This allows researchers to instantiate diverse planning architectures within a common computational framework that is tailored towards speech communication research. Existing DNF frameworks, such as COSIVINA \cite{schoener-etal2016} and its Python port \cite{schneegans2021}, are general-purpose simulators, whereas PyPhonPlan directly integrates DNFs with task dynamic models of articulatory control, thus providing a unified planning-to-production framework. The major contributions of this paper are:

\begin{enumerate}
    \item An open-source Python toolkit for simulating phonetic planning using coupled DNFs and task dynamics.
    \item A modular architecture for flexible model development and customization, including multi-layer fields, coupling mechanisms, memory fields, and tract variable simulations.
    \item Examples and code notebooks to promote reproducibility, extensibility, and cumulative computational research.
\end{enumerate}

The rest of this paper outlines the model components, before illustrating its functionality via a highly-simplified three-layer perception/planning/memory model of interactive speech. All software and examples are available on GitHub at:\\\url{https://github.com/samkirkham/PyPhonPlan}.

\section{Model architecture and implementation}

In this section, we review the components of the dynamic field model implemented in PyPhonPlan. For an excellent and extensive tutorial introduction to dynamic neural fields see \cite{schoener-etal2016}.

\subsection{Field dynamics}

The dynamics of a one-dimensional articulatory or acoustic control variable can be formalized as a dynamic neural field \cite{erlhagen-schoener2002}, with activation across a neural population defined over the metric range of $x$,

\begin{multline}
\tau \dot{u}(x,t) = -u(x,t) + h + s(x,t)
\\
+ \int k(x-x') g(u(x',t))dx' + q\xi(x,t),
\label{dnf}
\end{multline}

where $\tau$ dictates the rate of field evolution and $u(x,t)$ is time-dependent activation at each field site $x$. The negative sign $-u(x,t)$ ensures stable decay towards the resting level $h$. $s(x,t)$ represents sensory input to the field, and $\xi(x,t)$ is Gaussian noise scaled by a coefficient $q$ \cite{schoener-etal2016}. The interaction kernel $k(x - x')$ defines excitatory and inhibitory forces across the DNF, which is explained in Section \ref{interactions}.

\begin{figure}
  \centering
  \includegraphics[width=\columnwidth]{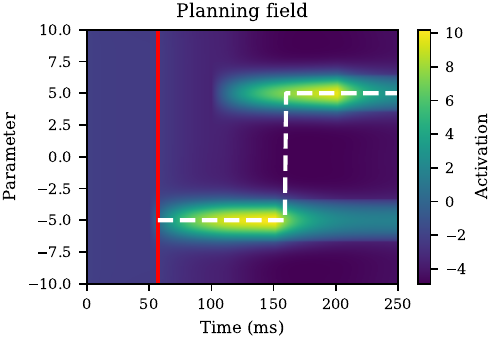}
  \caption{Heatmap of the planning-layer of a DNF with two competing inputs at $x = -5$ and $x = +5$. The red line indicates the start of above-threshold activation; the dotted white line tracks the location of above-threshold peak activation.}
  \label{fig:dnf}
\end{figure}

\subsection{Inputs}

The field in (\ref{dnf}) has its own intrinsic dynamics, but without sensory input the field's dynamics will converge to a sub-threshold resting level. Inputs to the DNF drive activation away from the resting level and excite localized areas of the field. Inputs can represent task-related input (i.e.~responding to an experimental prompt or cue), retrieval of a speech planning representation from memory, or sensory/perceptual inputs (i.e.~speech perception). An input $s(x,t)$ is defined as a distribution over a parameter $x$ with amplitude $a$, centroid $p$ and width $w$,

\begin{equation}
s(x,t) = \sum_{i} a_{i} \exp \left[ - \frac{(x-p_{i})^2}{2w^2_{i}} \right].
\label{dnf_input}
\end{equation}

This defines gestural inputs as inherently distributional, in contrast to the invariant target values in many task dynamic implementations. Inputs are typically time-bound; in the current implementation, timing is manually stipulated, but we discuss more principled timing mechanisms in Section \ref{discussion}. Figure \ref{fig:dnf} visualizes the dynamics of a single DNF, where two time-scheduled inputs at $x = -5$ and $x = +5$ drive activation above the resting level at different field locations.

\subsection{Field interactions and thresholds}
\label{interactions}

A key property of DNFs is that field sites interact with one another:~nearby sites excite each other while distant sites inhibit each other. This interaction profile has two key consequences:~it stabilizes peaks of activation once they form (through local self-excitation), and it enables competition between peaks, where activation of one peak will inhibit others. These dynamics are captured by the interaction kernel $k(x-x')$, which takes a `Mexican hat' profile, with local excitation $c_{\text{excite}}$ surrounded by lateral inhibition $c_{\text{inhibit}}$, plus a global inhibition term $c_{\text{global}}$:

\begin{multline}
k(x - x') = \frac{c_{\text{excite}}}{\sqrt{2 \pi \sigma^2_{\text{excite}}}} \exp \left[ -\frac{(x-x')^2}{2 \sigma^2_{\text{excite}}} \right] 
\\
- \frac{c_{\text{inhibit}}}{\sqrt{2 \pi \sigma^2_{\text{inhibit}}}} \exp \left[ -\frac{(x-x')^2}{2 \sigma^2_{\text{inhibit}}} \right] - c_{\text{global}}.
\label{kernel}
\end{multline}

The interaction kernel is gated by a sigmoid function,

\begin{equation}
g(u) = \frac{1}{1+\exp (-\beta (u-\alpha))},
\label{dnf_sigmoid}
\end{equation}

where $\beta$ is the slope of the sigmoid and $\alpha$ is a threshold parameter (typically $\alpha = 0$). This ensures that only field sites with activation above $\alpha$ contribute to the interaction dynamics.

\subsection{Memory field}
\label{memory}

\begin{figure}
  \centering
  \includegraphics[width=\columnwidth]{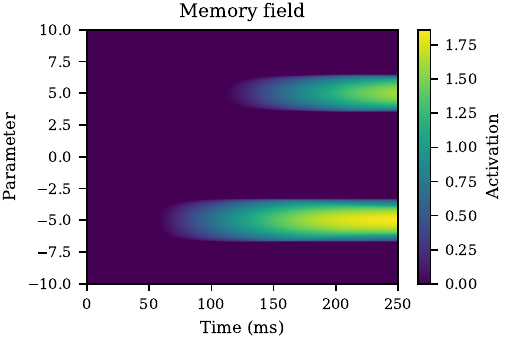}
  \caption{Memory field coupled to the planning field in Figure~\ref{fig:dnf}. The memory trace reflects a history of above-threshold planning activation, with a broader spatial profile due to convolution with $w(x-x')$.}
  \label{fig:memory}
\end{figure}

In addition to the standard fields described above, PyPhonPlan implements memory fields (see Figure \ref{fig:memory}), which are a special kind of field coupled to the planning field that stores a history of activation traces \cite{samuelson-etal2011}. This stored history can then bias the planning field on subsequent productions. Memory dynamics are implemented as a Hebbian layer over the memory field $u_{\text{mem}}(x,t)$ in (\ref{eq:memory}) subject to local interactions $w(x-x')$.

\begin{equation}
\dot{u}_{\text{mem}}(x,t)  =
\begin{cases} 
    \frac{1}{\tau_{\text{mem}}} [-u_{\text{mem}}(x,t) \\
    + \int w(x-x') g(u(x',t)) dx'], 
    & u(x,t) > \alpha \\[0.5em]
    \frac{1}{\tau_{\text{decay}}} [-u_{\text{mem}}(x,t)], 
    & u(x,t) \leq \alpha
\end{cases}.
\label{eq:memory}
\end{equation}

When the planning field  $u(x,t)$ exceeds the $\alpha$ threshold, the sigmoid $g(u)$ determines the pattern of activation that is convolved into the memory field, at a rate determined by $\tau_{\text{mem}}$. When $u(x,t) \leq \alpha$, memory decays at a rate of $\tau_{\text{decay}}$. The memory field evolves on a slower timescale  than the planning field, while memory decay is slowest, such that $\tau_{\text{decay}} > \tau_{\text{mem}} > \tau$.

\subsection{Cross-field coupling}
\label{coupling}

In multi-layer models, fields can be coupled with strength $c$, weighting the influence of perception or memory on planning. However, a strong perceptual input could push the planning field over the activation threshold, involuntarily triggering production even when no response input is present. PyPhonPlan avoids this via an (optional) latched gate on the planning field:

\begin{equation}
\gamma(t) =
\begin{cases}
    1 & \text{if } s_{\text{planning}}(x, t) > 0 \text{ for any } x, \\
    \gamma(t - \Delta t) & \text{if } u_{\text{planning}}(x, t) > \alpha \text{ for any } x, \\
    0 & \text{otherwise}
\end{cases}.
\label{eq:gating}
\end{equation}

When $\gamma(t) = 0$, activation is clamped so it cannot exceed threshold $\alpha$, ensuring that coupling from perception or memory can shape sub-threshold activation but cannot initiate production without an explicit input cue. The gate opens ($\gamma = 1$) when direct input arrives to the planning field ($s_{\text{planning}}(x, t) > 0$) and holds its value ($\gamma(t - \Delta t)$) while activation is above-threshold.
  
\subsection{Task dynamics}

We integrate DNF planning dynamics with a critically-damped harmonic oscillator model \cite{saltzman-munhall1989} for the evolution of tract variables, where $m = 1$, $k$ is a constant, and $b = 2\sqrt k$,

\begin{equation}
m\ddot{x} + b\dot{x} + k(x - x^*(t)) = 0.
\end{equation}

Following \cite{tilsen2019, kirkham-strycharczuk2024, shaw2025}, we derive the tract variable's time-varying target $x^*(t)$ from the planning DNF. The spatial position $x$ corresponding to peak activation in $u$ from the planning field at each time step is taken as the planned target value as in (\ref{eq:max}). An alternative formulation is to use a single constant value for $x^*(t)$, which can be derived from identifying the parameter value that corresponds to a peak activation plateau.

\begin{equation}
x^*(t)= \arg\max_x u(x,t).
\label{eq:max}
\end{equation}

In addition to the above core components, PyPhonPlan contains methods for visualization, animations, extracting DNF activation peaks, task dynamic solvers, and example simulation cases. See the GitHub repository for further information.

\section{Application: Coupled perception-planning-memory dynamics}

The PyPhonPlan GitHub repository contains fully-documented code notebooks for a range of DNF simulation cases. In this section, we present a three-layer model of phonetic shadowing \cite{goldinger1998}, which couples perception, planning and memory fields, and simulates tract variables from DNF activations. Parameter values and additional information can be found in the GitHub repository at \texttt{examples/shadowing.ipynb}.

\begin{figure*}[t]
  \centering
  \includegraphics[width=\textwidth]{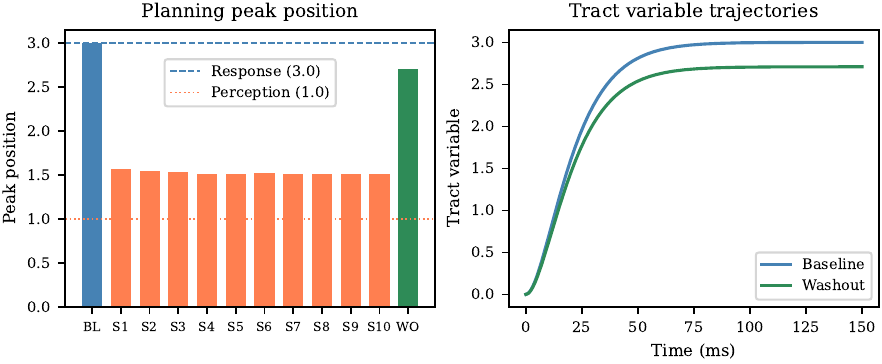}
  \caption{Shadowing simulation results. Left: planning peak position across baseline with no perceptual input (BL), 10 shadowing trials with strong perceptual input (S1--S10), and washout with no perceptual input (WO). Right: tract variable trajectories for baseline versus washout, showing subtle articulatory consequences of memory-driven convergence.}
  \label{fig:sim}
\end{figure*}

\subsection{Three-layer field}

We define a three-layer architecture of perception, planning and memory fields. For the sake of simplicity, we assume an arbitrary parameter space for $x \in [-10, 10]$, which is a hypothesized articulatory dimension. We first model the perception field as capturing the \textit{influence} of auditory perception on the planning field, rather than modelling the auditory-acoustic signal directly. The perception field takes the form

\begin{multline}
\tau_{\text{perception}}\dot{u}_{\text{perception}}(x,t)  = -u_{\text{perception}}(x,t) + h + s_{\text{perception}}(x,t)
\\
+ \int z(x-x') g(u_{\text{perception}}(x',t))dx' + q\xi(x,t),
\label{eq:perception}
\end{multline}

where $z(x-x')$ is a perception-specific interaction kernel. The advantage of a separate perception field, rather than perception as a simple input to the planning field, is that it allows perception to have its own intrinsic timescale and stabilization dynamics (e.g.~competition between perceptual inputs).

The memory field is implemented exactly as described in Section \ref{memory}, so Equation (\ref{eq:memory}) is not repeated here. The speech planning field also follows the generic DNF architecture,

\begin{multline}
\tau \dot{u}(x,t) = -u(x,t) + h + c_{\text{perception}} u_{\text{perception}}(x,t)
\\
+ c_{\text{memory}} u_{\text{memory}}(x,t) + c_{\text{response}}s_{\text{response}}(x,t)
\\
+ \int k(x-x') g(u(x',t))dx' + q\xi(x,t),
\label{eq:planning}
\end{multline}

but with coupling to $u_{\text{perception}}(x,t)$ and $u_{\text{memory}}(x,t)$ with strengths $c_{\text{perception}}$ and $c_{\text{memory}}$. These weights reflect the relative strength of perception/memory coupling, such as attentional strength or socially-modulated perception. In the following example, the interaction between production/perception fields is linear and unidirectional (perception $\rightarrow$ planning). The input $s_{\text{response}}(x,t)$ is a response input, prompted by retrieving an appropriate motor plan in response to a cue to speak. The $\gamma$-gating described in Section \ref{coupling} prevents perceptual coupling from involuntarily triggering production, but residual perception activation at response onset can bias the spatial location of peak formation. The memory field coupling $u_{\text{memory}}(x,t)$ means that the current state of memory additionally influences the planning field, while the memory field is simultaneously updated as a consequence of ongoing planning dynamics.

\subsection{Simulating phonetic shadowing}

We now simulate field dynamics during a simple shadowing task paradigm \cite{goldinger1998}, which takes the following form:

\begin{enumerate}
    \item \textbf{Baseline}:~speaker produces a word cued by an orthographic representation on a computer screen.
    \item \textbf{Shadow}:~speaker hears a (pre-recorded) talker producing the same word and is prompted to repeat the word they hear.
    \item \textbf{Washout}:~speaker again produces the same word, cued by an orthographic representation on a computer screen.
\end{enumerate}

Human speakers typically show phonetic convergence towards the interlocutor's voice during shadowing, with limited evidence for persistent post-shadowing convergence \cite{goldinger1998, babel2012}. In this sense, the shadowing task represents a highly constrained form of phonetic accommodation, whereby speakers adapt their speech in response to an interlocutor's speech. The simulations in this section follow the same structure as the above shadowing task paradigm; we simulate 1 baseline trial, 10 shadowing trials, and 1 washout trial. Note that $s_{\text{response}}(x,t) = 3$ is the speaker's intended production and $s_{\text{perception}}(x,t) = 1$ is the auditory prompt. As the intentional production and auditory prompt differ, this creates the opportunity for convergence effects during shadowing. During each trial, $u_{\text{memory}}(x,t)$ is updated based on the current activation profile in the planning field. This predicts that our model speaker should converge to the perceptual input, and the resulting memory field may show evidence of convergence during washout. Perception $t = [0,100]$ and response $t = [100, 200]$ inputs are timed sequentially, with the perception field time constant ($\tau = 10$) providing residual bias during planning peak formation.

\subsection{Simulation results}
Figure \ref{fig:sim} shows simulation results. The left panel shows the location of peak activation measured during the response across all trials. During baseline, the planning peak settles at the response position ($x = 3$). During shadowing, the auditory prompt ($x = 1$) induces a strong perception input that draws the planning peak toward the perceived token ($x = 1.56$ on S1), a bias that persists across shadowing trials. The washout planning peak retains some of this convergence effect, shifting $-0.29$ units from baseline ($x = 2.71$). This shift is driven by accumulated memory traces from shadowing trials, which bias the planning field. The right panel shows the corresponding tract variable trajectories for baseline and washout trials, with targets derived from the plateau peak positions. The washout trajectory reaches a lower target position than baseline, demonstrating how field-level convergence propagates to articulatory output. This is a simple, artificial example for the purpose of illustration; accommodation is rarely one-shot and of such large magnitude. Yet it illustrates how the model architecture predicts interactional phonetic phenomena as emergent from the coupled dynamics of speech planning, perception and memory.

\section{Discussion}
\label{discussion}

PyPhonPlan lowers the barrier to entry for dynamical models of phonetic planning, with a computational toolkit that is tailored to the requirements of researchers in speech communication. The flexible and modular architecture permits a range of simulation designs, with further examples in the GitHub repository. Several design choices leave room for further extension. One area for significant improvement is that input timing is currently stipulated. Future versions could implement coupled oscillator models \cite{nam-saltzman2003}, conditions of satisfaction for input deactivation \cite{sandamirskaya-schoner2010, chaturvedi-shaw2025b}, or competitive queuing for selection-activation dynamics \cite{tilsen2016, tilsen2022}. Second, the current task dynamic integration extracts time-varying targets from field peaks but lacks the full articulator-to-tract-variable mapping characteristic of standard models \cite{saltzman-munhall1989}. Third, the models shown here are 1D articulatory variables, but DNFs can scale to 2--4 dimensions for capturing multi-dimensional features (such as formants or x/y coordinates). However, DNFs do not scale well to much higher dimensions; while this encourages an interpretable and biophysically-grounded approach, it constrains modelling of high-dimensional problems and requires well-motivated theoretical commitments about which control variables to model.

A longer-term aim is to integrate PyPhonPlan with an extended Task Dynamic Application (TADA) architecture \cite{nam-etal2004}, enabling the use of gestural scores as inputs to the DNF and providing time-varying targets to the full TADA articulatory model. State feedback control is another important frontier; models such as FACTS \cite{parrell-etal2019} incorporate feedback into the task dynamic framework, while DNF \cite{chaturvedi-shaw2025} and other dynamical \cite{tilsen2022} architectures for feedback monitoring also show promise. Finally, the coupling of speaker-listener DNFs will facilitate new agent-based models of interactive speech dynamics, with the advantage of inherently dynamic and rich phonetic representations. For example, exemplar models of speech processing tend towards static phonetic representations, where memory retrieval and execution are instantaneous \cite{pierrehumbert2001}. In contrast, the DNF architecture explicitly models the dynamics of memory, perception and production, with dynamics specified at every scale.

\section{Conclusion}

In conclusion, PyPhonPlan is an open-source toolkit for building flexible models of phonetic planning using coupled dynamic neural fields and task dynamics. It provides a modular architecture that supports reproducible experimentation and transparent model development, thereby advancing a framework for cumulative computational research in speech communication.

\section{Acknowledgments}

S.K. was funded by AHRC fellowship AH/Y002822/1 and The Royal Society grant APX\textbackslash R1\textbackslash 251102, the latter of which was also supported by The Leverhulme Trust. 

\section{Generative AI Use Disclosure}

Anthropic's Claude Code (Opus 4.6) was used for debugging and automated testing of the accompanying software, with solutions subsequently verified by the author. No Generative AI was used in the writing of the manuscript.

\bibliographystyle{IEEEtran}
\bibliography{refs}

\end{document}